# Convex Sparse Spectral Clustering: Single-view to Multi-view

Canyi Lu, *Student Member, IEEE*, Shuicheng Yan, *Senior Member, IEEE*, and Zhouchen Lin, *Senior Member, IEEE*

**Spectral Clustering (SC) is one of the most widely used methods for data clustering. It first finds a low-dimensional embedding U of data by computing the eigenvectors of the normalized Laplacian matrix, and then performs k-means on $\mathbf{U}^\top$ to get the final clustering result. In this work, we observe that, in the ideal case, $\mathbf{UU}^\top$ should be block diagonal and thus sparse. Therefore we propose the Sparse Spectral Clustering (SSC) method which extends SC with sparse regularization on $\mathbf{UU}^\top$. To address the computational issue of the nonconvex SSC model, we propose a novel convex relaxation of SSC based on the convex hull of the fixed rank projection matrices. Then the convex SSC model can be efficiently solved by the Alternating Direction Method of Multipliers (ADMM). Furthermore, we propose the Pairwise Sparse Spectral Clustering (PSSC) which extends SSC to boost the clustering performance by using the multi-view information of data. Experimental comparisons with several baselines on real-world datasets testify to the efficacy of our proposed methods.**

## I. Introduction

Clustering algorithms are useful tools to explore data structures and have been employed in many disciplines. The objective of clustering is to partition a data set into groups (or clusters) such that the data points in the same cluster are more similar than those in other clusters. Many clustering algorithms have been developed with different motivations. Spectral Clustering (SC) [1] is one of the most widely used clustering methods with several variants being proposed [2], [3], [4], [5]. They have been applied for image segmentation [6], motion segmentation [7], co-clustering problems of words and documents [8], and genes and conditions [9]. This paper aims to improve the basic SC algorithm by utilizing a sparse regularizer.

### A. Related Work

A large body of work has been conducted on spectral clustering with focus on different aspects and applications [10],

This work was supported by the Singapore National Research Foundation under its International Research Centre, Singapore Funding Initiative and administered by the IDM Programme Office. The work of Z. Lin was supported in part by the National Basic Research Program of China (973 Program) under Grant 2015CB352502, in part by the National Natural Science Foundation (NSF) of China under Grant 61272341 and Grant 61231002, and in part by the Microsoft Research Asia Collaborative Research Program.

C. Lu and S. Yan are with the Department of Electrical and Computer Engineering, National University of Singapore, Singapore (e-mail: canyilu@gmail.com; eleyans@nus.edu.sg).

Z. Lin is with the Key Laboratory of Machine Perception (Ministry of Education), School of Electronics Engineering and Computer Science, Peking University, Beijing 100871, China, and also with the Cooperative Medianet Innovation Center, Shanghai Jiaotong University, Shanghai 200240, China (e-mail: zlin@pku.edu.cn).

[11], [12], [13]. Generally, existing approaches to improving spectral clustering performance can be categorized into two paradigms: (1) how to construct robust affinity matrix (or graphs) so as to improve the clustering performance by using the standard spectral algorithms [14], [15], [16], [17]; (2) how to improve the clustering result when the way of generating a data affinity matrix is fixed [18], [19], [20]. This paper is related to the second paradigm.

Before going on, we introduce our notation conventions first. We use boldface capital (e.g. $\mathbf{A}$), boldface lowercase (e.g. $\mathbf{a}$) and lowercase (e.g. $a$) symbols to represent matrices, vectors and scalars, respectively. In particular, $\mathbf{1}$ denotes the vector of all ones and $\mathbf{I}$ denotes the identity matrix. We use $\text{Tr}(\mathbf{A})$ to denote the trace of $\mathbf{A}$ and $\langle \mathbf{A}, \mathbf{B} \rangle$ to denote the inner product of $\mathbf{A}$ and $\mathbf{B}$. The set of symmetric matrices with size $n \times n$ is denoted as $\mathbb{S}^n$. For $\mathbf{A}, \mathbf{B} \in \mathbb{S}^n$, $\mathbf{A} \preceq \mathbf{B}$ or $\mathbf{B} \succeq \mathbf{A}$ means that $\mathbf{B} - \mathbf{A}$ is positive semi-definite, or $\mathbf{B} - \mathbf{A} \succeq \mathbf{0}$. For $\mathbf{a}, \mathbf{b} \in \mathbb{R}^n$, $\mathbf{a} \leq \mathbf{b}$ or $\mathbf{b} \geq \mathbf{a}$ means that each element of $\mathbf{b} - \mathbf{a}$ is nonnegative. We use $\|\mathbf{A}\|_0$, $\|\mathbf{A}\|_1$ and $\|\mathbf{A}\|$ to denote the $\ell_0$-norm (number of nonzero elements of $\mathbf{A}$), $\ell_1$-norm ($\sum_{ij} |a_{ij}|$) and Frobenius norm of $\mathbf{A}$, respectively.

Assume we are given the data matrix $\mathbf{X} = [\mathbf{X}_1, \cdots, \mathbf{X}_k] = [\mathbf{x}_1, \cdots, \mathbf{x}_n] \in \mathbb{R}^{d \times n}$, where $\mathbf{X}_i \in \mathbb{R}^{d \times n_i}$ denotes the data matrix belonging to group $i$, $d$ is the dimension, $n$ is the number of data points and $k$ is the number of clusters. SC [2] partitions these $n$ points into $k$ clusters as follows:

1. Compute the affinity matrix $\mathbf{W} \in \mathbb{R}^{n \times n}$ where each element $w_{ij}$ measures the similarity between $\mathbf{x}_i$ and $\mathbf{x}_j$;
2. Compute the normalized Laplacian matrix $\mathbf{L} = \mathbf{I} - \mathbf{D}^{-\frac{1}{2}} \mathbf{W} \mathbf{D}^{-\frac{1}{2}}$, where $\mathbf{D}$ is a diagonal matrix with each diagonal element $d_{ii} = \sum_{i=1}^{n} w_{ij}$;
3. Compute $\mathbf{U} \in \mathbb{R}^{n \times k}$ by solving

$$\min_{\mathbf{U} \in \mathbb{R}^{n \times k}} \langle \mathbf{UU}^\top, \mathbf{L} \rangle \quad \text{s.t.} \quad \mathbf{U}^\top \mathbf{U} = \mathbf{I}; \tag{1}$$

4. Form $\hat{\mathbf{U}} \in \mathbb{R}^{n \times k}$ by normalizing each row of $\mathbf{U}$ to have unit Euclidean length;
5. Treat each row of $\hat{\mathbf{U}}$ as a point in $\mathbb{R}^k$, and cluster them into $k$ groups by k-means.

The rows of $\mathbf{U}$ can be regarded as the low-dimensional embedding of $\mathbf{X}$. The columns of a solution $\mathbf{U}$ to (1) are the first $k$ eigenvectors of $\mathbf{L}$ corresponding to the smallest $k$ eigenvalues. Generally, there are three critical steps which affect the performance of spectral clustering: (1) the construction of the affinity matrix which measures the degree of similarity between data points; (2) the way to construct the Laplacian matrix $\mathbf{L}$ or normalization; (3) the way to find the low-dimensional embedding $\mathbf{U}$ of $\mathbf{X}$ based on $\mathbf{L}$. For the first step, the purpose of the construction of the affinity matrix is to model the neighborhood relationship between data points. There are several popular ways to construct the affinity matrix, such as the k-nearest neighbor graph and the fully connected graph. There are also many work focused on the second step. For example, Ratio-cut and Normalized-cut, aim at different normalizations on the Laplacian matrix $\mathbf{L}$ when given $\mathbf{W}$ [10], [6]. It was proved in [12] that the key difference between Ratio-cut and Normalized-cut is the error measure used to find the closest doubly stochastic approximation of the affinity matrix during the normalization step. Zass et al. [21] proposed to find the optimal doubly stochastic matrix under the Frobenius norm. But it neglected the positive semidefinite constraint during the normalization step which makes the approximation to the affinity matrix less accurate. The work [19] further addressed this issue and proposed an efficient solver. Different from most of previous work which focused on the first two steps, this work focuses on the third step which aims to find the low-dimensional embedding $\mathbf{U}$ of $\mathbf{X}$ in a new way.

As will be detailed discussed in Section II, we observe that $\mathbf{U}\mathbf{U}^\top$ is block diagonal when $\mathbf{W}$ (or $\mathbf{L}$) is block diagonal in the ideal case. However, such a structure of $\mathbf{U}\mathbf{U}^\top$ is not explicitly used in previous work. Note that a block diagonal matrix is also sparse. Motivated by this observation, we propose the Sparse Spectral Clustering (SSC) method to find a better $\mathbf{U}$ with the sparse regularization on $\mathbf{U}\mathbf{U}^\top$. This makes SSC own a better interpretation of the affinity matrix $|\mathbf{U}\mathbf{U}^\top|$ than SC. However, the original formulation of SSC is nonconvex and thus is challenging to solve. We propose a convex relaxation of SSC based on a convex hull, called the Fantope [22], which provides a tight relaxation of jointly fixed rank and orthogonality constraint on the positive semidefinite cone. Then we propose to solve the convex SSC problem by Alternating Direction Method of Multipliers (ADMM) [23] which guarantees to find the optimal solution.

Furthermore, we extend our SSC for data clustering by using the multi-view information. It is observed that the real-world data usually have multiple representations or views. These views provide complementary information to each other. For example, given an image, one can represent it by different kinds of features, e.g., colors, textures and bag-of-word. They describe the image in different ways or views. Using the multi-view information generally boost the performance on the learning task.

Assume that we are given data which have $m$ views. Let $\mathbf{W}_i$ and $\mathbf{L}_i$ denote the affinity matrix and the normalized Laplacian matrix of the data in the $i$-th view. The recent work [24] proposed the Pairwise Spectral Clustering (PSC) which computes $\mathbf{U}_i$'s by solving

$$\min_{\{\mathbf{U}_i\}} \sum_{i=1}^m \langle \mathbf{U}_i\mathbf{U}_i^\top, \mathbf{L}_i \rangle + \frac{\alpha}{2} \sum_{\substack{1 \le i,j \le m \\ i \ne j}} \|\mathbf{U}_i\mathbf{U}_i^\top - \mathbf{U}_j\mathbf{U}_j^\top\|^2$$
$$\text{s.t.} \quad \mathbf{U}_i^\top \mathbf{U}_i = \mathbf{I}, \ \mathbf{U}_i \in \mathbb{R}^{n \times k}, \ i = 1, \cdots, m, \quad (2)$$

where $\alpha > 0$ is a trade-off parameter. The second term of the above objective encourages the pairwise similarities of examples under the new representation to be similar across all the views. So the objective of PSC trades off the SC objectives and the spectral embedding disagreement term. A main issue of PSC is that problem (2) is nonconvex and the optimal solution is not computable. The work [24] used an alternating optimization method to solve (2), i.e., updating each $\mathbf{U}_i$ by fixing all the other variables. However, such a method has no convergence guarantee when $m > 2$. The nonconvexity of (2) may limit the application of PSC. There are some works which improve the efficiency of PSC, e.g., the work [25] relaxes the constraint (2) and solves all $\mathbf{U}_i$'s simultaneously.

Similar to SSC, we propose the Pairwise Sparse Spectral Clustering (PSSC) method which further encourages each $\mathbf{U}_i\mathbf{U}_i^\top$ to be sparse. It is originally nonconvex, and we propose a convex formulation of PSSC with a Fantope constraint. Finally ADMM is applied to compute the globally optimal solution to the convex PSSC problem.

### B. Contributions

We summarize our contributions of this work as follows:

1. We propose a convex Sparse Spectral Clustering (SSC) model which encourages $\mathbf{U}\mathbf{U}^\top$ to be sparse. This improves the interpretation of SC since $|\mathbf{U}\mathbf{U}^\top|$ can be regarded as a new affinity matrix which is expected to be block diagonal in the ideal case.
2. We propose the Pairwise Sparse Spectral Clustering (PSSC) method which seeks to improve the clustering performance by leveraging the multi-view information. Note that our PSSC is convex and thus its optimal solutions is computable. This addresses the computational issue in PSC whose objective is nonconvex.
3. We present an efficient ADMM algorithm to solve the convex SSC and PSSC problems.

The remainder of this paper is organized as follows. Section II and III present our SSC and PSSC models, respectively. Section IV provides the optimization details for the PSSC problem. The experiments are reported in Section V, and Section VI concludes this work.

## II. SPARSE SPECTRAL CLUSTERING

This section presents the Sparse Spectral Clustering (SSC) algorithm for single-view data clustering. Assume that we are given the affinity matrix $\mathbf{W}$ and normalized Laplacian matrix $\mathbf{L}$ as in SC. Let us consider the solution $\mathbf{U}$ to (1). It has an interesting connection with the indicator matrix $\mathbf{C} \in \mathbb{R}^{n \times k}$, whose row entries indicate to which group the points belong. That is, if $\mathbf{x}_i$ belongs to the group $l$, $c_{il} = 1$ and $c_{ij} = 0$ for all $j \ne l$. Consider the ideal case that the affinity matrix $\mathbf{W}$ is block diagonal, i.e., $w_{ij} = 0$ if $\mathbf{x}_i$ and $\mathbf{x}_j$ are in different clusters. Then we have

$$\hat{\mathbf{U}} = \mathbf{C}\mathbf{R},$$

where $\mathbf{R} \in \mathbb{R}^{k \times k}$ can be any orthogonal matrix [2]. In this case,

$$\hat{\mathbf{U}}\hat{\mathbf{U}}^\top = \mathbf{C}\mathbf{C}^\top = \begin{bmatrix} \mathbf{1}_{n_1}\mathbf{1}_{n_1}^\top & \mathbf{0} & \cdots & \mathbf{0} \\ \mathbf{0} & \mathbf{1}_{n_2}\mathbf{1}_{n_2}^\top & \cdots & \mathbf{0} \\ \vdots & \vdots & \ddots & \vdots \\ \mathbf{0} & \mathbf{0} & \cdots & \mathbf{1}_{n_k}\mathbf{1}_{n_k}^\top \end{bmatrix},$$

is block diagonal. So $\hat{\mathbf{U}}\hat{\mathbf{U}}^\top$ implies the true membership of the data clusters and it is naturally *sparse*. The sparse property also holds for $\mathbf{U}\mathbf{U}^\top$. However, in real-world applications, the formed affinity matrix $\mathbf{W}$ is usually not block diagonal, and thus $\mathbf{U}\mathbf{U}^\top$ may not be block diagonal for the solution $\mathbf{U}$ to (1) (actually $\mathbf{U}\mathbf{U}^\top$ is usually not sparse in the real applications). It can be seen that the quality of $\mathbf{U}\mathbf{U}^\top$ implicates the discriminability of the affinity matrix $\mathbf{W}$. Or $|\mathbf{U}\mathbf{U}^\top|$ can be regarded as a new affinity matrix implied by $\mathbf{W}$, which is is ideally block diagonal. The sparse subspace clustering method [7] approximated such a block diagonal structure by sparse coding. It was also verified in practice that the sparse regularizer is effective even $\mathbf{W}$ is not exactly block diagonal. This motivates our SSC model which encourages $\mathbf{U}\mathbf{U}^\top$ to be sparse in SC:

$$\min_{\mathbf{U} \in \mathbb{R}^{n \times k}} \langle \mathbf{U}\mathbf{U}^\top, \mathbf{L} \rangle + \beta \|\mathbf{U}\mathbf{U}^\top\|_0, \quad \text{s.t.} \quad \mathbf{U}^\top \mathbf{U} = \mathbf{I}, \quad (3)$$

where $\beta > 0$ trades off the objective of SC and the sparsity of $\mathbf{U}\mathbf{U}^\top$. For the affinity matrix $\mathbf{W}$, it is reasonable to assume that the intra-cluster connections are relatively strong while the inter-cluster connections are relatively weak. In this case, the elements of $\mathbf{U}\mathbf{U}^\top$ corresponding to the weak inter-cluster connections tends to be zeros, while the ones corresponding to the strong intra-cluster connections will be kept. So it is expected that that the clustering performance of SC can be improved by SSC by using the sparse regularization. Note that the way of using sparse regularization to approximate a block diagonal matrix is also used in [7], [11]. However, the key difference is that their methods aim to find a sparse affinity matrix while our SSC aims to find the low-dimensional embedding $\mathbf{U}$ when the affinity matrix is given.

Problem (3) is nonconvex and challenging to solve. We propose to convert (3) into a convex formulation by relaxing its feasible domain into a convex set. First, it is known that the $\ell_1$-norm is the convex envelope of $\ell_0$-norm within the $\ell_1$-ball. So we replace the $\ell_0$-norm as the $\ell_1$-norm on $\mathbf{U}\mathbf{U}^\top$. Second, for the nonconvex constraint consisting of all the fixed rank projection matrices, i.e., $\{\mathbf{U}\mathbf{U}^\top | \mathbf{U}^\top \mathbf{U} = \mathbf{I}\}$, we replace it with its convex hull.

**Theorem 1.** *[26], [27] Let $S_1 = \{\mathbf{U}\mathbf{U}^\top | \mathbf{U} \in \mathbb{R}^{n \times k}, \mathbf{U}^\top \mathbf{U} = \mathbf{I}\}$ and $S_2 = \{\mathbf{P} \in \mathbb{S}^{n \times n} | \mathbf{0} \preceq \mathbf{P} \preceq \mathbf{I}, Tr(\mathbf{P}) = k\}$. Then $S_2$ is the convex hull of $S_1$ and $S_1$ is exactly the set of extreme points of $S_2$.*

The convex body $S_2$ is also called the Fantope [22]. Theorem 1 is interesting and important in this work. Now, we give a simple proof which is helpful for understanding this result. It is easy to see that any convex combination of elements of $S_1$ lies in $S_2$. Also, using the spectral decomposition of $\mathbf{P}$, which has eigenvalues lying between 0 and 1 that sum to $k$, it is clear that any element of $S_2$ with rank greater than $k$ is not an extreme point. So the only candidates for extreme points are those with rank $k$, i.e., the elements of $S_2$. But it is not possible that some rank $k$ elements are extreme points and others are not, since the definition of $S_2$ does not in any way distinguish between different rank $k$ elements. Since a compact convex set must have extreme points, and is in fact the convex hull of its extreme points. The proof is completed.

By using Theorem 1, we now give a convex formulation of the nonconvex problem (3). We replace $\mathbf{U}\mathbf{U}^\top$ in (3) as $\mathbf{P}$, and relax the (feasible) problem domain to a convex set based on the relationship between $S_1$ and $S_2$ presented above; this leads to a convex formulation of SSC defined as follows

$$\min_{\mathbf{P} \in \mathbb{R}^{n \times n}} \langle \mathbf{P}, \mathbf{L} \rangle + \beta \|\mathbf{P}\|_1, \quad \text{s.t.} \quad \mathbf{0} \preceq \mathbf{P} \preceq \mathbf{I}, \ \text{Tr}(\mathbf{P}) = k. \quad (4)$$

It is interesting to see that (4) is equivalent to (1) without the sparsity term, or $\beta = 0$. In this case, $\mathbf{P} = \mathbf{U}\mathbf{U}^\top$ is optimal to (4) when $\mathbf{U}$ is optimal to (1) (see page 310 in [22]). If $\beta > 0$, the optimal solution to (4) is not guaranteed to be an extreme point of $S_2$. So problem (4) is a convex relaxtion of (3).

The optimal solution to (4) can be efficiently computed by the standard Alternating Direction Method of Multipliers (ADMM) [23]. Since (4) is a special case of (6) shown later, we only give the optimization details for (6) in Section IV.

After solving (4) with the solution $\mathbf{P}$, the solution $\mathbf{U}$ to (3) can be approximated by using the first $k$ eigenvectors corresponding to the largest $k$ eigenvalues of $\mathbf{P}$. Finally, we can obtain the clustering results as SC based on the rows of $\mathbf{U}$. See Algorithm 1 for the whole procedure of SSC.

It is worth mentioning that the work [28] proposes a sparse PCA program, which has a similar formulation of our (4). However, it is very different from our work. First, the task and physical meaning of sparse PCA is very different from our work which focuses on data clustering. For example, given the affinity matrix $\mathbf{W}$, $\mathbf{U}\mathbf{U}^\top$ can be interpreted as the refined affinity matrix in SC and SSC. This is quite different from sparse PCA which aims to find sparse principal components. Second, the $\ell_1$-norm in (4) of our SSC is a convex relaxation of $\ell_0$-norm in (3). Both the $\ell_1$-norm and $\ell_0$-norm encourage the refined affinity matrix to be sparse. However, the $\ell_1$-norm in (1) of [28] is used as a convex relaxation of $\ell_{2,0}$-norm which towards row sparsity. So though the relaxed formulations, our model (4) and model (1) in (Vu et al. 2013), look similar, their original or non-relaxed formulations are very different. This also reflects the key different purposes of sparse PCA and our SSC.

## III. MULTI-VIEW EXTENSION: PAIRWISE SPARSE SPECTRAL CLUSTERING

SC and our SSC can only use the single view information for data clustering. In this section, we present the Pairwise Sparse Spectral Clustering (PSSC) method which boosts the clustering performance by using the multi-view information. Assume that we are given data which have $m$ views and $\mathbf{L}_i$, $i = 1, \cdots, m$, are the corresponding normalized Laplacian matrices. Our PSSC model extends PSC in (2) by further

**Algorithm 1** Sparse Spectral Clustering (SSC)

**Input:** data matrix $\mathbf{X} \in \mathbb{R}^{d \times n}$, number of clusters $k$.
1. Compute the affinity matrix $\mathbf{W}$.
2. Compute the normalized Laplacian matrix $\mathbf{L}$ as in SC.
3. Compute $\mathbf{P}$ by solving (4).
4. Form $\mathbf{U}$ with its columns being the first $k$ eigenvectors corresponding to the largest $k$ eigenvalues of $\mathbf{P}$.
5. Form $\hat{\mathbf{U}} \in \mathbb{R}^{n \times k}$ by normalizing each row of $\mathbf{U}$ to have unit Euclidean length.
6. Treat each row of $\hat{\mathbf{U}}$ as a point in $\mathbb{R}^k$, cluster them into $k$ groups by k-means.

encouraging each $\mathbf{U}_i \mathbf{U}_i^\top$ to be sparse. The PSSC model is as follows:

$$\min_{\{\mathbf{U}_i\}} \sum_{i=1}^m \left( \langle \mathbf{U}_i \mathbf{U}_i^\top, \mathbf{L}_i \rangle + \beta \|\mathbf{U}_i \mathbf{U}_i^\top\|_0 \right) \\ + \frac{\alpha}{2} \sum_{\substack{1 \le i,j \le m \\ i \ne j}} \|\mathbf{U}_i \mathbf{U}_i^\top - \mathbf{U}_j \mathbf{U}_j^\top\|^2, \quad (5)$$

$$\text{s.t.} \quad \mathbf{U}_i^\top \mathbf{U}_i = \mathbf{I}, \ \mathbf{U}_i \in \mathbb{R}^{n \times k}, i = 1, \cdots, m,$$

where $\beta > 0$ trades off the objective of PSC and the sparse regularizers on $\mathbf{U}_i \mathbf{U}_i^\top$'s. For the simplicity, we use a common $\alpha$ for all pairwise co-regularizers and a common $\beta$ for the sparse regularizers on all views. Problem (5) is nonconvex and is challenging to solve. Similar to the convex formulation of SSC in (4), we relax (5) to a convex formulation as follows

$$\min_{\{\mathbf{P}_i\}} \sum_{i=1}^m (\langle \mathbf{P}_i, \mathbf{L}_i \rangle + \beta \|\mathbf{P}_i\|_1) + \frac{\alpha}{2} \sum_{\substack{1 \le i,j \le m \\ i \ne j}}^m \|\mathbf{P}_i - \mathbf{P}_j\|^2, \quad (6)$$

$$\text{s.t.} \quad \mathbf{0} \preceq \mathbf{P}_i \preceq \mathbf{I}, \ \text{Tr}(\mathbf{P}_i) = k, \mathbf{P}_i \in \mathbb{R}^{n \times n}, \ i = 1, \cdots, m.$$

Now the above problem is convex and the optimal solution can be computed efficiently. After solving (6) with $\mathbf{P}_i$'s, we compute each $\mathbf{U}_i$ by using the first $k$ eigenvectors corresponding to the largest $k$ eigenvalues of $\mathbf{P}_i$. Then $\mathbf{U}_i$'s can be combined (e.g., via column-wise concatenation) before running k-means [24].

Note that the work [24] also proposed a centroid based co-regularized SC model for multi-view data clustering. It can be also extended with sparse regularizers as our PSSC. However, since it was not always shown to be superior to PSC, we do not consider its extension in this work.

## IV. OPTIMIZATION BY ADMM

This section gives the optimization details for the proposed SSC in (4) and PSSC in (6). Note that the SSC problem (4) is only a special case of PSSC problem (6) by taking $m = 1$ and $\alpha = 0$. So we only give the optimization details of (6) by the standard ADMM.

The objective of (6) is non-separable. We first reformulate it as an equivalent formulation with separable objective by introducing some auxiliary variables

$$\min_{\{\mathbf{P}_i, \mathbf{Q}_i\}} \sum_{i=1}^m (\langle \mathbf{P}_i, \mathbf{L}_i \rangle + \beta \|\mathbf{P}_i\|_1) + \frac{\alpha}{2} \sum_{\substack{1 \le i,j \le m \\ i \ne j}} \|\mathbf{Q}_i - \mathbf{P}_j\|^2,$$

$$\text{s.t.} \quad \mathbf{0} \preceq \mathbf{Q}_i \preceq \mathbf{I}, \text{Tr}(\mathbf{Q}_i) = k, \quad (7)$$
$$\mathbf{P}_i = \mathbf{Q}_i, \ \mathbf{P}_i \in \mathbb{R}^{n \times n}, \ i = 1, \cdots, m.$$

The partial augmented Lagrangian function of (7) is

$$\mathcal{L}(\mathbf{P}_i, \mathbf{Q}_i, \mathbf{Y}_i) = \sum_{i=1}^m \left( \langle \mathbf{P}_i, \mathbf{L}_i \rangle + \beta \|\mathbf{P}_i\|_1 + \langle \mathbf{Y}_i, \mathbf{P}_i - \mathbf{Q}_i \rangle \right)$$

$$+ \frac{\mu}{2} \sum_{i=1}^m \|\mathbf{P}_i - \mathbf{Q}_i\|^2 + \frac{\alpha}{2} \sum_{\substack{1 \le i,j \le m \\ i \ne j}} \|\mathbf{Q}_i - \mathbf{P}_j\|^2,$$

where $\mathbf{Y}_i$'s are the dual variables and $\mu > 0$. By regarding $\{\mathbf{P}_i, i = 1, \cdots, m\}$ and $\{\mathbf{Q}_i, i = 1, \cdots, m\}$ as two blocks. Now we show how to update these two blocks alternately.

1. Fix others and minimize $\mathcal{L}$ w.r.t. $\{\mathbf{P}_i, i = 1, \cdots, m\}$. Note that $\mathcal{L}$ is separable for all $\mathbf{P}_i$'s. So each $\mathbf{P}_i$ can be updated independently by solving

$$\min_{\mathbf{P}_i} \langle \mathbf{P}_i, \mathbf{L}_i + \mathbf{Y}_i \rangle + \beta \|\mathbf{P}_i\|_1 + \frac{\mu}{2} \|\mathbf{P}_i - \mathbf{Q}_i\|^2 \\ + \frac{\alpha}{2} \sum_{\substack{1 \le j \le m \\ j \ne i}} \|\mathbf{Q}_j - \mathbf{P}_i\|^2.$$

Or equivalently

$$\min_{\mathbf{P}_i} \frac{\beta}{\alpha(m-1)+\mu} \|\mathbf{P}_i\|_1 \\ + \frac{1}{2} \left\| \mathbf{P}_i - \frac{\alpha \sum_{\substack{1 \le j \le m \\ j \ne i}} \mathbf{Q}_j + \mu \mathbf{Q}_i - \mathbf{L}_i - \mathbf{Y}_i}{\alpha(m-1)+\mu} \right\|^2. \quad (8)$$

Solving the above problem requires computing the proximal mapping of the $\ell_1$-norm, i.e.,

$$\min_{\mathbf{X}} \varepsilon \|\mathbf{X}\|_1 + \frac{1}{2} \|\mathbf{X} - \mathbf{B}\|^2,$$

which can be obtained by performing

$$\mathcal{S}(x) = \begin{cases} x - \varepsilon, & x > \varepsilon, \\ x + \varepsilon, & x < -\varepsilon, \\ 0, & \text{otherwise.} \end{cases}$$

on each element of $\mathbf{X}$.

2. Fix others and minimize $\mathcal{L}$ w.r.t. $\{\mathbf{Q}_i, i = 1, \cdots, m\}$. Note that $\mathcal{L}$ is separable for all $\mathbf{Q}_i$'s. So each $\mathbf{Q}_i$ can be updated independently by solving

$$\min_{\mathbf{Q}_i} \langle -\mathbf{Y}_i, \mathbf{Q}_i \rangle + \frac{\mu}{2} \|\mathbf{P}_i - \mathbf{Q}_i\|^2 + \frac{\alpha}{2} \sum_{\substack{1 \le j \le m \\ j \ne i}} \|\mathbf{Q}_i - \mathbf{P}_j\|^2$$

$$\text{s.t.} \quad \mathbf{0} \preceq \mathbf{Q}_i \preceq \mathbf{I}, \ \text{Tr}(\mathbf{Q}_i) = k.$$

**Algorithm 2** Solve Problem (6) by ADMM
**Input:** $\mathbf{L}_i, i = 1, \cdots, m, \alpha, \beta, k$.
**Initialize:** $\mathbf{P}_i, \mathbf{Q}_i, \mathbf{Y}_i, i = 1, \cdots, m, \rho > 1, \mu, \mu_{\max}$.
**Output:** $\mathbf{P}_i^*, i = 1, \cdots, m$.
**while** not converged **do**

1. For each $i$, fix others and update $\mathbf{P}_i$ by solving (8).
2. For each $i$, fix others and update $\mathbf{Q}_i$ by solving (9).
3. For each $i$, update $\mathbf{Y}_i$'s by (12).
4. Update $\mu$ by $\mu = \min(\rho\mu, \mu_{\max})$.

**end while**

Or equivalently

$$\min_{\mathbf{Q}_i} \frac{1}{2} \left\| \mathbf{Q}_i - \frac{\alpha \sum_{\substack{1 \leq j \leq m \\ j \neq i}} \mathbf{P}_j + \mu \mathbf{P}_i + \mathbf{Y}_i}{\alpha(m-1) + \mu} \right\|^2 \quad (9)$$

s.t. $\mathbf{0} \preceq \mathbf{Q}_i \preceq \mathbf{I}, \; \mathrm{Tr}(\mathbf{Q}_i) = k$.

Solving (9) requires computing a proximal projection onto the convex set $\{\mathbf{Q}_i | \mathbf{0} \preceq \mathbf{Q}_i \preceq \mathbf{I}, \mathrm{Tr}(\mathbf{Q}_i) = k\}$. It is equivalent to solving a simple quadratic programming.

**Theorem 2.** *Given any square matrix* $\mathbf{A} \in \mathbb{R}^{n \times n}$, *let* $\mathbf{B} = (\mathbf{A} + \mathbf{A}^\top)/2$ *and* $\mathbf{B} = \mathbf{U}Diag(\boldsymbol{\lambda})\mathbf{U}^\top$ *be the spectral decomposition of the symmetric matrix* $\mathbf{B}$. *Then the solution to the following problem*

$$\min_{\mathbf{Q}} \frac{1}{2} \|\mathbf{Q} - \mathbf{A}\|^2, \; s.t. \; \mathbf{0} \preceq \mathbf{Q} \preceq \mathbf{I}, \; Tr(\mathbf{Q}) = k, \quad (10)$$

*is* $\mathbf{Q}^* = \mathbf{U}Diag(\boldsymbol{\rho}^*)\mathbf{U}^\top$, *where* $\boldsymbol{\rho}^*$ *is the solution to*

$$\min_{\boldsymbol{\rho}} \frac{1}{2} \|\boldsymbol{\rho} - \boldsymbol{\lambda}\|^2, \; s.t. \; \mathbf{0} \leq \boldsymbol{\rho} \leq \mathbf{1}, \; \boldsymbol{\rho}^\top \mathbf{1} = k. \quad (11)$$

The proof of Theorem 2 can be found in the Appendix. It can be seen that solving (9) reduces to solving (11). Problem (11) is called the capped simplex projection problem which has been efficiently solved in [29].

3. Update the dual variables by

$$\mathbf{Y}_i = \mathbf{Y}_i + \mu(\mathbf{P}_i - \mathbf{Q}_i), \; i = 1, \cdots, m. \quad (12)$$

The whole procedure of ADMM for solving (6) can be found in Algorithm 2. Note that problem (6) is convex. The obtained solution by ADMM for convex problem is guaranteed to be optimal with the convergence rate $O(1/K)$ [30], where $K$ is the number of iterations. The computational complexity of Algorithm 2 is $O(Kmn^3)$, where $m$ is the number of views and $n$ is the number of data samples. In the real applications, $m \ll n$. In the experiments, we find that $K$ is around $100 \sim 200$. Compared with $O(n^3)$ by the traditional SC, the computational complexity of Algorithm 2 is $K$ times higher. So the advantage in the computational cost of SC over the Algorithm 2 is obvious only when $n$ is relatively small. Though developing the scalable SSC algorithm is not the main concern of this work, it will be an interesting future work.

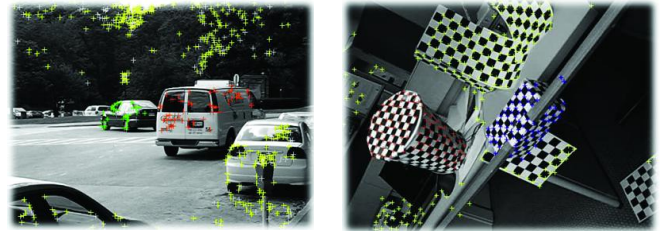

(a)

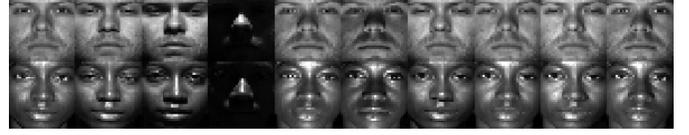

(b)

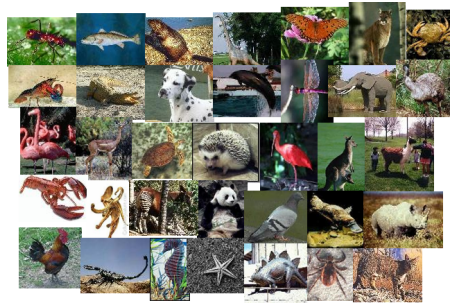 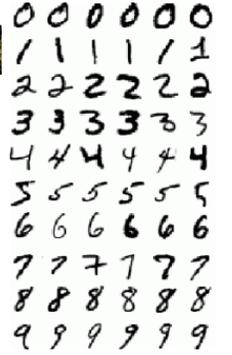

(c) (d)

Fig. 1: Examples of the datasets: (a) Hopkins 155; (b) Extended Yale B (c) Caltech-101; (d) UCI Digit.

## V. EXPERIMENTS

In this section, we conduct extensive experiments on several real-world datasets to demonstrate the effectiveness of our proposed SSC and PSSC for both single view and multiview data clustering. The experiments are divided into two parts. The first part is to demonstrate the superiority of SSC over SC for single-view data clustering, while the second part is to show the effectiveness of PSSC for multi-view data clustering. We also report the running time of the compared methods on several datasets and exam the sensitiveness of parameter choices in our methods.

### A. Single-view Experiments

To show the effectiveness of our sparse regularizer, we first test SSC on the single-view clustering problems, including motion segmentation on the Hopkins 155 dataset[1] and face clustering on the Extended Yale B dataset [31]. To apply SC and SSC, one first needs to construct the affinity matrix. We compute the affinity matrix by sparse subspace clustering [7], [32] and low-rank representation [33] (denoted as $\ell_1$-graph and LRR-graph respectively in Table I and Table II) which are two state-of-the-art methods. The codes of $\ell_1$-graph and LRR-graph with their default settings from the authors' homepages are used.

---
[1] http://www.vision.jhu.edu/data/hopkins155/

TABLE I: Segmentation errors (%) on the Hopkins 155 dataset. For SSC in this table, we set $\beta = 10^{-5}$.

|  | Affinity matrix | $\ell_1$-graph | | LRR-graph | |
|---|---|---|---|---|---|
|  | Method | SC | SSC | SC | SSC |
| 2 Motions | Mean | 1.83 | **1.81** | **1.33** | 1.36 |
|  | Median | 0.00 | 0.00 | 0.00 | 0.00 |
| 3 Motions | Mean | 4.40 | **4.35** | 2.51 | **1.65** |
|  | Median | 0.56 | 0.56 | 0.00 | 0.89 |
| All | Mean | 2.41 | **2.39** | 1.60 | **1.42** |
|  | Median | 0.00 | 0.00 | 0.00 | 0.00 |

TABLE II: Clustering errors (%) on the Extended Yale B dataset.

| Affinity matrix | $\ell_1$-graph | | LRR-graph | |
|---|---|---|---|---|
| Method | SC | SSC ($\beta = 10^{-4}$) | SC | SSC ($\beta = 10^{-3}$) |
| 5 Subjects | | | | |
| Mean | **6.28** | 6.56 | 25.16 | **21.27** |
| STD | 5.41 | 5.90 | 14.26 | 11.88 |
| 8 subjects | | | | |
| Mean | 8.81 | **4.99** | 35.96 | **28.85** |
| STD | 5.99 | 3.96 | 5.56 | 9.81 |
| 10 Subjects | | | | |
| Mean | 10.06 | **4.62** | 35.42 | **31.26** |
| STD | 4.96 | 2.57 | 0.99 | 4.01 |

The Hopkins 155 dataset consists of 155 video sequences, where 120 of the videos have two motions and 35 of the videos have three motions. On average, in the dataset, each sequence of 2 motions has $N = 266$ feature trajectories and $F = 30$ frames, while each sequence of 3 motions has $N = 398$ feature trajectories and $F = 29$ frames. Some example frames of this dataset can be found in Figure 1 (a). For each sequence, there are 39~550 data points drawn from two or three motions (a motion corresponds to a subspace). Each sequence is a sole data set and so there are 156 subspace segmentation problems in total. For each sequence, we use PCA to reduce the dimension (see [7] for the details) and then apply SC and SSC on the constructed affinity matrix by $\ell_1$-graph and LRR-graph. The mean and median of the error rates are reported in Table I.

The Extended Yale B dataset consists of 2,414 face images of 38 subjects. Each subject has 64 faces. Some example face images can be found in Figure 1 (b). We resize the images into $32 \times 32$ pixels and treat each 1,024-dimensional vectorized image as a data point. We construct three subsets which consist of all the images of the randomly selected 5, 8 and 10 subjects of this dataset. For each trial, we follow the settings in [7] to construct the affinity matrix by $\ell_1$-graph and LRR-graph and apply SC and SSC to achieve the clustering results. The experiments are repeated 20 times and the mean and standard deviation of the clustering error rates are reported in Table II.

From Table I and II, we have the following observations:

- For motion segmentation on the Hopkins 155 dataset, the clustering errors of SC and SSC are all small, and the improvement of SSC over SC is relatively limited. Note that the data in Hopkins 155 has small noises and the PCA projection preprocessing further reduces the noises [7], [33]. The affinity matrices learned by both $\ell_1$-graph and LRR are very discriminative. Thus, SC performs very well on this dataset (the segmentation errors on most of the sequences are zero). SSC further improves the performance of SC, but the improvement on the mean of all 155 sequences is relatively limited.
- For face clustering on the Extended Yale B dataset, the clustering errors of SC and SSC are relatively larger than the results in the Hopkins 155 dataset. The main reason is that this dataset is heavily corrupted due to the large variation of the illumination. Note that the raw pixels of the images are used without preprocessing. Thus, SC does not perform very well on this dataset, and SSC significantly outperforms SC in most cases. This verifies the effectiveness of our proposed sparse regularizer on $\mathbf{UU}^\top$. Note that the best results of SSC are achieved when $\beta = 10^{-4}$ for the $\ell_1$-graph and $\beta = 10^{-3}$ for the LRR-graph. A possible reason is that the $\ell_1$-graph tends to be more sparse than the LRR-graph, and thus it will be more necessary to enforce the sparsity on the LRR-graph in our SSC.

### B. Multi-view Experiments

This subsection conducts several experiments to demonstrate the effectiveness of PSSC for multi-view data clustering.

#### 1) Datasets

We report experimental results on five real-world datasets which are widely used for the multi-view data clustering [24], [34]. The statistics of these datasets are summarized in Table IV. We give a brief description of each dataset as follows.

- **Caltech-101** dataset [35] contains 8,677 images of objects belonging to 101 categories. Some example images can be found in Figure 1 (c). This dataset is challenging for clustering. We use a subset which has 75 samples with 5 underlying clusters. We choose the "pixel features", "Pyramid Histogram Of Gradients", and "Sparse Localized Features" as our three views to represent each image.
- **3-sources** dataset[2] was collected from three well known online news sources: BBC, Reuters and Guardian. It consists of 416 distinct news manually categorized into six classes. Among them, 169 are reported in all three sources with each source serving as one independent view of a story. The word frequency is the feature used to describe stories for all the three views.
- **Reuters** dataset [36] contains feature characteristics of documents originally written in five different languages and their translations, over a common set of 6 categories. We use documents in English as the first view and their translations to other four languages as another four views. We randomly sample 600 documents from this collection in a balanced manner, with each of the 6 clusters containing 100 documents.
- **WebKB** dataset[3] consists of webpages collected from four universities: Texas, Cornell, Washington and Wisconsin, and each webpage can be described by the content view and the link view.
- **UCI Digit** dataset[4] consists of handwritten numerals ('0'–'9') extracted from a collection of Dutch utility maps. Some example images can be found in Figure 1 (d). This dataset consists of 2,000 samples with 200 in each category, and it is represented in terms of six features. We choose the 76 Fourier coefficients of the character shapes, 216 profile correlations and 64 Karhunen-Love coefficients as three views to represent each image.

---
[2]http://mlg.ucd.ie/datasets/3sources.html

TABLE IV: Statistics of the real-world multi-view datasets

| dataset | samples | views | clusters |
|---|---|---|---|
| Caltech 101 | 75 | 3 | 5 |
| 3-sources | 169 | 3 | 6 |
| Reuters | 600 | 5 | 6 |
| WebKB | 1051 | 2 | 2 |
| UCI Digit | 2000 | 3 | 10 |

*2) Compared Methods and Settings*

The following methods will be used for comparison. The first three ones are single view methods while the others are multi-view methods.

- **SC [2]:** Perform the traditional SC on the most informative view, i.e., the view that achieves the best performance.
- **SSC:** Perform our SSC on the most informative view.
- **LD-SSC [19]:** Perform LD-SSC on the most informative view.
- **Feature Concatenation:** Concatenate the features of each view, and then run SC using the normalized Laplacian matrix derived from the joint view representation of the data.
- **Kernel Addition:** Combine different kernels by adding them, and then run SC on the corresponding normalized Laplacian matrix.
- **Robust Multi-View Spectral Clustering (RMSC) [34]:** A recent multi-view clustering work based on the low-rank and sparse decomposition.
- **PSC [24]:** Extend SC to use multi-view information.
- **PSSC:** Our extension of PSC with sparse regularization.

For the performance evaluation, we use five metrics to measure the clustering performances: precision, recall, F-score, normalized mutual information (NMI) and adjusted rand index(Adj-RI) [37] as that in [34]. For these measures, the higher values indicate better clustering performances.

Note that all the compared methods shown above call k-means at the final step to get the clustering results. The clustering results by k-means with different initializations may be different. So we run k-means 20 times and report the means and standard deviations of the performance measures.

[3] http://www.cs.cmu.edu/afs/cs/project/theo-20/www/data/
[4] http://archive.ics.uci.edu/ml/datasets/Multiple+Features

In all experiments, we use the Gaussian kernel to compute the affinity similarities. The standard deviation of the kernel is taken equal to the median of the pair-wise Euclidean distances between the data points.

In this work, we assume that the number of clusters $k$ is known. So SC, LD-SSC, Feature Concatenation and Kernel Addition have no parameter. For $\alpha$ in PSC, we vary it from 0.01 to 0.05 and the best result is reported. RMLS has a parameter and we use the default setting in [34]. For our SSC and PSSC, we tune the parameters and report the best results. For $\beta$ in SSC (4), we choose it from $\{10^{-3}, 10^{-4}, 10^{-5}\}$. For $\alpha$ and $\beta$ in PSSC (6), we choose $\alpha$ from $\{10^{-1}, 10^{-2}\}$ and $\beta$ from $\{10^{-3}, 10^{-4}, 10^{-5}\}$.

*3) Results*

The clustering results on five real-world datasets are shown in Table III. For each dataset, the results by SC, LD-SSC and SSC are the best performances which use the single view information of data. The other four methods all use the multi-view information of data. The best results are shown in bold. Note that for the Caltech-101 dataset, we do not report the result of Feature Concatenation since only the affinity matrices of all views are known. For the WebKB dataset, the standard deviations in parentheses are not 0 but very small. We only keep the first three digits after the decimal point.

From Table III, we have the following observations:
- In most cases, our PSSC achieves the best performance or it is comparable to the best result. PSSC is more stable than other multi-view methods which do not perform well on some datasets, e.g., RMSC on the 3-sources dataset. The improvement of PSSC over other methods partially supports our sparse regularizations on $\mathbf{U}_i\mathbf{U}_i^\top$'s.
- Comparing the results of SC and SSC, it can be seen that SSC always outperforms SC. Note that the optimal solutions to SC in (1) (though nonconvex) and SSC in (4) are computable. SSC is more general than SC due to the additional sparse regularization on $\mathbf{U}\mathbf{U}^\top$. Such a result verifies the effectiveness of the sparse regularization on $\mathbf{U}\mathbf{U}^\top$. Also it can be seen that PSSC outperforms PSC similarly. To see the reason more intuitively, based on the UCI Digit dataset, we plot the affinity matrix of the first view, $\mathbf{U}\mathbf{U}^\top$ by SC, $\mathbf{P}$ by SSC, $\sum_i |\mathbf{U}_i\mathbf{U}_i^\top|$ by PSC and $\sum_i |\mathbf{P}_i|$ by PSSC in Figure 2. The matrices (using their absolute values if necessary) by the compared four methods play a similar role, i.e., they can be regarded as affinity matrices. It can be seen that the block diagonal structures of our SSC and PSSC are more salient than the non-sparse SC and PSC, and thus their performance is expected to be better. However, notice that it does not mean that the sparser solutions of SC and PSSC always lead to better performance. The larger $\beta$'s in (4) and (6) lead to sparser solutions, but $\beta$'s should not be too large since the discriminant information mainly comes from the first term of (4) and (6) and they should not be ignored.
- For the results on the first three datasets, the single view methods, SC and SSC, perform even better than some multi-view clustering methods, e.g., Feature Concatena-tion and Kernel Addition. It is expected that the multi-view clustering performance is better than the single view, when the views of data are discriminative and diverse. But note that this may not be guaranteed in practice. More importantly, the way for using the multi-view information of data is crucial.

TABLE III: Comparison results on five datasets. On each dataset, 20 test runs with different random initializations of k-means are conducted and the average performance and the standard deviation (numbers in parentheses) are reported.

| Dataset | Method | F-score | Precision | Recall | NMI | Adj-RI |
|---|---|---|---|---|---|---|
| Caltech-101 | SC | 0.504 (0.032) | 0.481 (0.028) | 0.529 (0.036) | 0.503 (0.036) | 0.381 (0.039) |
| | LD-SSC | 0.533 (0.022) | 0.488 (0.032) | 0.549 (0.032) | 0.521 (0.028) | 0.412 (0.035) |
| | SSC | 0.546 (0.027) | 0.505 (0.025) | **0.595 (0.029)** | **0.578 (0.049)** | 0.430 (0.034) |
| | Kernel Addition | 0.463 (0.032) | 0.436 (0.047) | 0.497 (0.031) | 0.439 (0.040) | 0.327 (0.046) |
| | RMSC | 0.536 (0.007) | 0.521 (0.001) | 0.551 (0.014) | 0.519 (0.007) | 0.423 (0.007) |
| | PSC | 0.534 (0.029) | 0.513 (0.034) | 0.556 (0.027) | 0.524 (0.027) | 0.419 (0.038) |
| | PSSC | **0.558 (0.021)** | **0.531 (0.020)** | 0.588 (0.025) | 0.559 (0.025) | **0.448 (0.026)** |
| 3-sources | SC | 0.502 (0.030) | 0.535 (0.037) | 0.475 (0.047) | 0.473 (0.028) | 0.362 (0.035) |
| | LD-SSC | 0.521 (0.045) | 0.536 (0.027) | 0.502 (0.034) | 0.482 (0.025) | 0.393 (0.038) |
| | SSC | 0.538 (0.046) | 0.532 (0.021) | **0.548 (0.080)** | 0.481 (0.023) | 0.400 (0.050) |
| | Feature Concat. | 0.497 (0.041) | 0.547 (0.064) | 0.457 (0.031) | 0.520 (0.031) | 0.361 (0.057) |
| | Kernel Addition | 0.473 (0.030) | 0.534 (0.026) | 0.425 (0.033) | 0.452 (0.017) | 0.337 (0.035) |
| | RMSC | 0.395 (0.024) | 0.468 (0.031) | 0.342 (0.021) | 0.374 (0.023) | 0.248 (0.031) |
| | PSC | 0.546 (0.052) | 0.604 (0.049) | 0.498(0.056) | **0.566 (0.028)** | 0.426 (0.063) |
| | PSSC | **0.568 (0.037)** | **0.616 (0.036)** | 0.528 (0.043) | 0.551 (0.025) | **0.450 (0.044)** |
| Reuters | SC | 0.371 (0.012) | 0.342 (0.008) | 0.405 (0.024) | 0.319 (0.013) | 0.233 (0.012) |
| | LD-SSC | 0.377 (0.019) | 0.348 (0.015) | 0.409 (0.030) | 0.331 (0.025) | 0.241 (0.017) |
| | SSC | 0.384 (0.017) | **0.354 (0.013)** | 0.419 (0.034) | **0.342 (0.027)** | **0.249 (0.018)** |
| | Feature Concat. | 0.369 (0.013) | 0.332 (0.014) | 0.416 (0.021) | 0.320 (0.018) | 0.227 (0.016) |
| | Kernel Addition | 0.370 (0.016) | 0.338 (0.018) | 0.410 (0.022) | 0.309 (0.023) | 0.231 (0.021) |
| | RMSC | 0.373 (0.013) | 0.345 (0.014) | 0.407 (0.016) | 0.325 (0.016) | 0.237 (0.017) |
| | PSC | 0.372 (0.017) | 0.339 (0.016) | 0.414 (0.029) | 0.319 (0.032) | 0.233 (0.019) |
| | PSSC | **0.388 (0.009)** | 0.345 (0.022) | **0.455 (0.027)** | 0.331 (0.017) | 0.248 (0.017) |
| WebKB | SC | 0.889 (0.000) | 0.824 (0.000) | 0.965 (0.000) | 0.532 (0.000) | 0.618 (0.000) |
| | LD-SSC | 0.901 (0.000) | 0.840 (0.000) | 0.967 (0.000) | 0.579 (0.000) | 0.671 (0.000) |
| | SSC | 0.902 (0.000) | 0.846 (0.000) | **0.967 (0.000)** | 0.582 (0.000) | 0.673 (0.000) |
| | Feature Concat. | 0.947 (0.000) | 0.947 (0.000) | 0.947 (0.000) | 0.718 (0.000) | 0.845 (0.000) |
| | Kernel Addition | 0.947 (0.000) | 0.947 (0.000) | 0.947 (0.000) | 0.718 (0.000) | 0.845 (0.000) |
| | RMSC | 0.956 (0.000) | 0.962 (0.000) | 0.951 (0.000) | 0.761 (0.000) | 0.873 (0.000) |
| | PSC | 0.948 (0.000) | 0.956 (0.000) | 0.940 (0.000) | 0.729 (0.000) | 0.850 (0.000) |
| | PSSC | **0.957 (0.000)** | **0.965 (0.000)** | 0.950 (0.000) | **0.769 (0.000)** | **0.878 (0.000)** |
| UCI Digit | SC | 0.640 (0.032) | 0.580 (0.047) | 0.716 (0.016) | 0.710 (0.019) | 0.596 (0.038) |
| | LD-SSC | 0.645 (0.031) | 0.591 (0.039) | 0.717 (0.018) | 0.715 (0.021) | 0.611 (0.035) |
| | SSC | 0.661 (0.028) | 0.617 (0.042) | 0.714 (0.016) | 0.721 (0.018) | 0.621 (0.033) |
| | Feature Concat. | 0.456 (0.015) | 0.443 (0.019) | 0.469 (0.013) | 0.560 (0.016) | 0.394 (0.018) |
| | Kernel Addition | 0.746 (0.026) | 0.729 (0.040) | 0.764 (0.017) | 0.783 (0.013) | 0.717 (0.030) |
| | RMSC | 0.813 (0.043) | 0.780 (0.056) | 0.826 (0.029) | 0.834 (0.021) | 0.791 (0.048) |
| | PSC | 0.757 (0.056) | 0.732 (0.073) | 0.786 (0.039) | 0.796 (0.030) | 0.729 (0.064) |
| | PSSC | **0.826 (0.044)** | **0.792 (0.069)** | **0.864 (0.015)** | **0.849 (0.022)** | **0.805 (0.050)** |

- Despite the effectiveness, SSC and PSSC have higher computational cost. Figure 3 plots the running times of SC, SSC, RMSC, PSC and PSSC on four datasets[5]. It can be seen that SC is more efficient than SSC. Though PSC is faster than PSSC, it is superiority is not that significant as SC over SSC, since PSC requires several iterations to achieve a local solution. Also, the new parameter $\beta$ in our SSC and PSSC may leads to additional effort for parameter tuning. As shown in Figure 4, we found that

---

[5]The running times of Feature Concat. and Kernel Addition are not plotted since they are very similar to SC. The running time on Caltech-101 dataset is not reported since it is small.

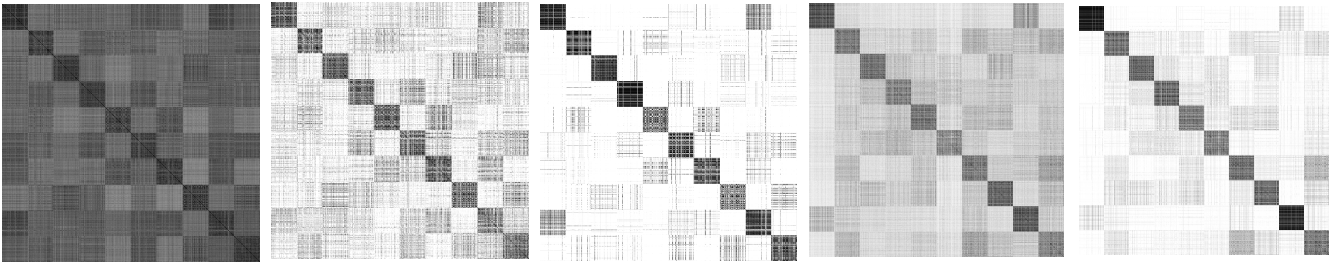

(a) affinity matrix $\mathbf{W}_1$    (b) $\mathbf{U}\mathbf{U}^\top$ by SC    (c) $\mathbf{P}$ by SSC    (d) $\sum_i |\mathbf{U}_i\mathbf{U}_i^\top|$ by PSC    (e) $\sum_i |\mathbf{P}_i|$ by PSSC

Fig. 2: Plots of (a) the affinity matrix which uses the first view information of the UCI Digit dataset, (b) $\mathbf{U}\mathbf{U}^\top$ where $\mathbf{U}$ is obtained by SC in (1), (c) $\mathbf{P}$ by SSC in (4), (d) $\sum_i |\mathbf{U}_i\mathbf{U}_i^\top|$ by PSC in (2) and (e) $\sum_i |\mathbf{P}_i|$ by PSSC in (6).

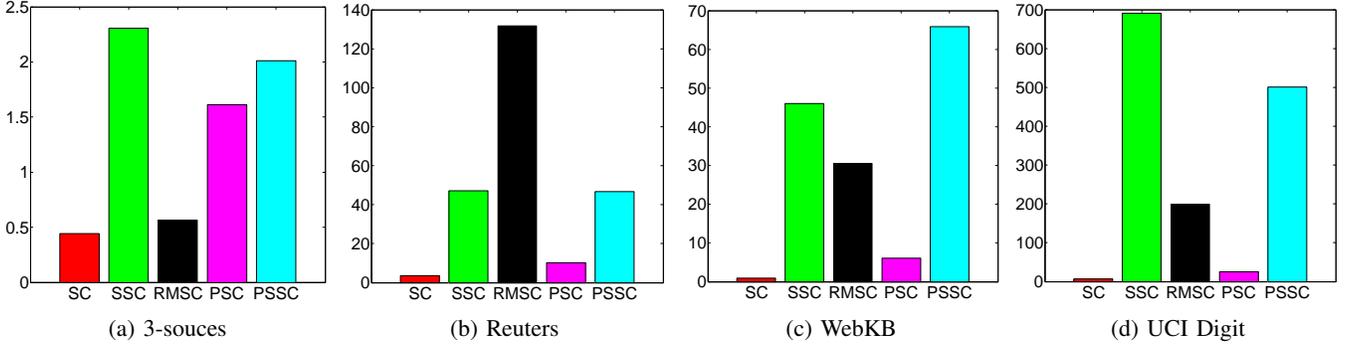

(a) 3-souces    (b) Reuters    (c) WebKB    (d) UCI Digit

Fig. 3: Running times (in seconds) of some compared methods on four datasets.

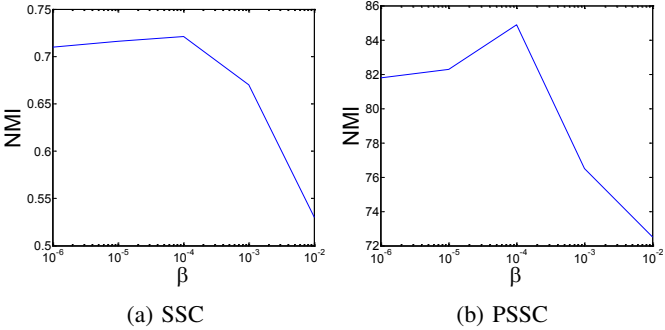

(a) SSC    (b) PSSC

Fig. 4: Plots of NMI v.s. $\beta$ in SSC and PSSC ($\alpha = 0.01$) on the UCI Digit dataset.

SSC and PSSC perform well and stably when $\beta \leq 10^{-3}$.

## VI. CONCLUSIONS AND FUTURE WORK

This paper proposed the Sparse Spectral Clustering (SSC) model which extended the traditional spectral clustering method with a sparse regularization. The original formulation of SSC is nonconvex and we presented a tight relaxation based on the convex hull of fixed rank projection matrices. We also proposed the convex Pairwise Sparse Spectral Clustering (PSSC) model which further improved SSC by exploiting multi-view information of data. Extensive experiments conducted on several real-world datasets demonstrated the effectiveness of our methods.

There remain some problems for future exploration. First, though SSC in (4) looks reasonable, we are lack of proof that it will actually compute a reasonable clustering. Under what conditions can it be expected to do well? Second, problem (4) is a convex relaxation of (3). How tight the relaxation is? Under what conditions can we recover the solution of (3) by solving (4)? We may see the answers to these questions in the future.

## APPENDIX

**Proof of Theorem 2.** Note that $\mathbf{Q}$ is symmetric. This implies that $\|\mathbf{Q}-\mathbf{A}\|^2 = \|\mathbf{Q}-\mathbf{A}^\top\|^2$. Thus

$$\begin{aligned}\frac{1}{2}\|\mathbf{Q}-\mathbf{A}\|^2 &= \frac{1}{4}\|\mathbf{Q}-\mathbf{A}\|^2 + \frac{1}{4}\|\mathbf{Q}-\mathbf{A}^\top\|^2 \\ &= \frac{1}{2}\left\|\mathbf{Q}-(\mathbf{A}+\mathbf{A}^\top)/2\right\|^2 + c(\mathbf{A}) \\ &= \frac{1}{2}\|\mathbf{Q}-\mathbf{B}\|^2 + c(\mathbf{A}),\end{aligned}$$

where $c(\mathbf{A})$ depends only on $\mathbf{A}$. Hence (10) is equivalent to

$$\min_{\mathbf{Q}} \frac{1}{2}\|\mathbf{Q}-\mathbf{B}\|^2, \text{ s.t. } \mathbf{0} \preceq \mathbf{Q} \preceq \mathbf{I}, \text{ Tr}(\mathbf{Q})=k. \quad (13)$$

Both $\mathbf{Q}$ and $\mathbf{B}$ are symmetric. Let $\rho_1 \geq \rho_2 \geq \cdots \geq \rho_n$ and $\lambda_1 \geq \lambda_2 \geq \cdots \geq \lambda_n$ be the ordered eigenvalues of $\mathbf{Q}$ and $\mathbf{B}$, respectively. By using the fact $\text{Tr}(\mathbf{Q}^\top\mathbf{B}) \leq \sum_{i=1}^n \rho_i \lambda_i$ in [38], we have

$$\begin{aligned}\|\mathbf{Q}-\mathbf{B}\|_F^2 &= \text{Tr}(\mathbf{Q}^\top\mathbf{Q}) - 2\text{Tr}(\mathbf{Q}^\top\mathbf{B}) + \text{Tr}(\mathbf{B}^\top\mathbf{B}) \\ &= \sum_{i=1}^m \rho_i^2 - 2\text{Tr}(\mathbf{Q}^\top\mathbf{B}) + \sum_{i=1}^m \lambda_i^2 \\ &\geq \sum_{i=1}^m \left(\rho_i^2 - 2\rho_i\lambda_i + \lambda_i^2\right) \\ &= \|\boldsymbol{\rho}-\boldsymbol{\lambda}\|^2.\end{aligned}$$

It is easy to verify that the above equality holds when $\mathbf{Q}$ admits the spectral decomposition $\mathbf{Q} = \mathbf{U}\text{Diag}(\boldsymbol{\rho})\mathbf{U}^\top$. On the other hand, the constraints $\mathbf{0} \preceq \mathbf{Q} \preceq \mathbf{I}$ and $\text{Tr}(\mathbf{Q}) = k$ are equivalent to $\mathbf{0} \leq \boldsymbol{\rho} \leq \mathbf{1}$ and $\boldsymbol{\rho}^\top \mathbf{1} = k$, respectively. Thus $\mathbf{Q}^* = \mathbf{U}\text{Diag}(\boldsymbol{\rho}^*)\mathbf{U}^\top$ is optimal to (13) with $\boldsymbol{\rho}^*$ being optimal to (11). The proof is completed. ∎

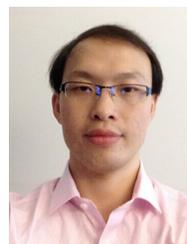

**Canyi Lu** received the bachelor degree in mathematics from the Fuzhou University in 2009, and the master degree in the pattern recognition and intelligent system from the University of Science and Technology of China in 2012. He is currently a Ph.D. student with the Department of Electrical and Computer Engineering at the National University of Singapore. His current research interests include computer vision, machine learning, pattern recognition and optimization. He was the winner of the Microsoft Research Asia Fellowship 2014. His homepage is https://sites.google.com/site/canyilu/.

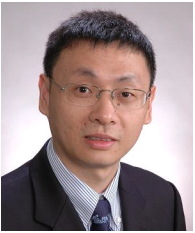

**Shuicheng Yan** is currently an Associate Professor at the Department of Electrical and Computer Engineering at National University of Singapore, and the founding lead of the Learning and Vision Research Group (http://www.lv-nus.org). Dr. Yan's research areas include machine learning, computer vision and multimedia, and he has authored/co-authored hundreds of technical papers over a wide range of research topics, with Google Scholar citation >27,000 times and H-index 68. He is ISI Highly-cited Researcher, 2014 and IAPR Fellow 2014. He has been serving as an associate editor of IEEE TKDE, TCSVT and ACM Transactions on Intelligent Systems and Technology (ACM TIST). He received the Best Paper Awards from ACM MM'13 (Best Paper and Best Student Paper), ACM MM12 (Best Demo), PCM'11, ACM MM10, ICME10 and ICIMCS'09, the runner-up prize of ILSVRC'13, the winner prize of ILSVRC14 detection task, the winner prizes of the classification task in PASCAL VOC 2010-2012, the winner prize of the segmentation task in PASCAL VOC 2012, the honourable mention prize of the detection task in PASCAL VOC'10, 2010 TCSVT Best Associate Editor (BAE) Award, 2010 Young Faculty Research Award, 2011 Singapore Young Scientist Award, and 2012 NUS Young Researcher Award.

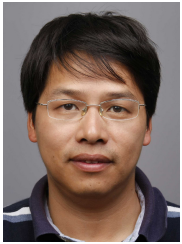

**Zhouchen Lin** received the Ph.D. degree in Applied Mathematics from Peking University, in 2000. He is currently a Professor at Key Laboratory of Machine Perception (MOE), School of Electronics Engineering and Computer Science, Peking University. He is also a Chair Professor at Northeast Normal University and a Guest Professor at Beijing Jiaotong University. Before March 2012, he was a Lead Researcher at Visual Computing Group, Microsoft Research Asia. He was a Guest Professor at Shanghai Jiaotong University and Southeast University, and a Guest Researcher at Institute of Computing Technology, Chinese Academy of Sciences. His research interests include computer vision, image processing, computer graphics, machine learning, pattern recognition, and numerical computation and optimization. He is an Associate Editor of IEEE Trans. Pattern Analysis and Machine Intelligence and International J. Computer Vision, an area chair of CVPR 2014, ICCV 2015, NIPS 2015 and AAAI 2016, and a Senior Member of the IEEE.